\documentclass[manuscript,screen,nonacm]{acmart}
\AtBeginDocument{%
  \providecommand\BibTeX{{%
    \normalfont B\kern-0.5em{\scshape i\kern-0.25em b}\kern-0.8em\TeX}}}

\usepackage{float}

\begin{document}

\title{Translation-based Video-to-Video Synthesis}

\author{Pratim Saha}

\email{psaha@uab.edu}
\orcid{1234-5678-9012}

\affiliation{%
  \institution{The University of Alabama at Birmingham}
  \streetaddress{10th Ave S}
  \city{Birmingham}
  \state{Alabama}
  \country{USA}
  \postcode{35294}
}

\author{Chengcui Zhang}
\affiliation{%
  \institution{The University of Alabama at Birmingham}
  \streetaddress{10th Ave S}
  \city{Birmingham}
  \country{USA}}
\email{czhang02@uab.edu}

\renewcommand{\shortauthors}{Saha and Zhang}

\begin{abstract}
Translation-based Video Synthesis (TVS) has emerged as a vital research area in computer vision, aiming to facilitate the transformation of videos between distinct domains while preserving both temporal continuity and underlying content features. This technique has found wide-ranging applications, encompassing video super-resolution, colorization, segmentation, and more, by extending the capabilities of traditional image-to-image translation to the temporal domain. One of the principal challenges faced in TVS is the inherent risk of introducing flickering artifacts and inconsistencies between frames during the synthesis process. This is particularly challenging due to the necessity of ensuring smooth and coherent transitions between video frames. Efforts to tackle this challenge have induced the creation of diverse strategies and algorithms aimed at mitigating these unwanted consequences. This comprehensive review extensively examines the latest progress in the realm of TVS. It thoroughly investigates emerging methodologies, shedding light on the fundamental concepts and mechanisms utilized for proficient video synthesis. This survey also illuminates their inherent strengths, limitations, appropriate applications, and potential avenues for future development.
\end{abstract}




\maketitle

\section{Introduction}
Deep learning has made tremendous strides in the last several years for analyzing and comprehending digital images for many computer vision applications. Many of such applications involve video synthesis, e.g., transforming a black-white movie into a color movie, or creating a more artistic, animation-like version of a real-world video. Research of this kind is generally referred to as ``translation-based video synthesis (TVS)''. The  general objective of TVS is to change the external style of a frame while keeping its internal content intact. It takes a frame $X_A$ from a source video in domain A and transforms it into a frame $X_B$ in domain B, ensuring the content remains the same while the style aligns with that of domain B.
While substantial recent works on image-to-image translation (i2i) techniques have produced astounding results \cite{alotaibi2020deep,kaji2019overview,chen2021overview,kamil2019literature}, less exploration has been seen for TVS. For a variety of tasks such as video colorization \cite{zhang2019deep}, medical imaging \cite{nie2016estimating}, model-based reinforcement learning \cite{arulkumaran2017brief}, and computer graphics rendering \cite{kajiya1986rendering}, the capacity to combine dynamic visual representations is crucial. The TVS translation task is more difficult than the i2i task. In addition to the constraints for frame-wise realism and semantic preservation that are also present in the i2i translation, TVS approaches also need to consider the temporal consistency for creating sequence-based results. One easy way to apply current i2i approaches for TVS is to treat video frames as distinct images and train an i2i model. Unfortunately, image-based generators do not have the capacity to make frame-wise realistic and temporally consistent translated videos. An important feature of TVS that previous i2i models struggled to achieve is the temporal consistency between video frames \cite{bansal2018recycle}. Moreover, semantic inconsistency, which refers to changes in the semantic labels throughout the unstable translation process, is another prevalent issue strongly connected with temporal inconsistency. In our best knowledge, there is no comprehensive survey in the literature that discusses in detail about recent advances in TVS. In this survey, we will cover state-of-the-art (SOTA) methods that concentrate on TVS for diverse tasks.

Addressing the issues inherent in i2i approaches for TVS, the introduction of video-to-video (v2v) translation by Wang et al. \cite{wang2018video} marked a significant step forward. Their approach integrate a spatio-temporal objective function within a well-structured conditional-GAN framework, successfully mitigating the flickering effect prevalent in i2i models for TVS. Building on this foundation, subsequent efforts by \cite{wei2018video,mallya2020world} introduced a residual error-based mechanism, incorporating both global and local consistency alongside conditional-GAN optimization. However, a substantial training dataset with ground labels is necessary for such methods. This limitation was addressed by Few-shot vid2vid \cite{wang2019few}, which not only reduced model dependency on extensive datasets but also enhanced generalization capacity. 

Further refining TVS methods, Bashkirova et al. \cite{bashkirova2018unsupervised} introduced a spatio-temporal 3D translator-based on i2i networks, although it suffered from a lack of regularization in spatio-temporal consistency. To combat this issue, Recycle-GAN \cite{bansal2018recycle} incorporated temporal factors into a Cycle-GAN framework \cite{zhu2017unpaired}, resulting in accurate prediction of future frames. Liu et al. in \cite{liu2020unsupervised} enhance the Recycle-GAN model by adding a tendency-invariant loss, which improves the modifications in generated frames and overall performance. However, the effectiveness of regularization in spatio-temporal consistency hinged on accurate motion estimation, a limitation addressed by capturing motion changes between consecutive frames \cite{chen2019mocycle, park2019preserving}. Mocycle-GAN \cite{chen2019mocycle} and STC-V2V \cite{park2019preserving} capitalized on motion translation via optical flow and image warping algorithms, enhancing temporal continuity in video synthesis. Despite their potential, these methods faced issues due to capacity constraints and domain gap issues, adversely affecting motion estimation and overall stability.
To overcome these challenges, Wang et al. \cite{wang2022learning} proposed synthetic optical flow to ensure consistent motion representation between future frames across domains, mitigating incorrect motion estimation. Leveraging Recurrent Neural Networks (RNNs) \cite{liu2021unsupervised, tulyakov2018mocogan} for temporal coherence, forward and backward RNN units gathered temporal and spatial information for current frames. However, to integrate the spatial and temporal context of the present frame, this technique considers all previous and subsequent frames, which is computationally intensive and challenging to implement in real-time scenarios.
All these models require complex architectures for preserving spatio-temporal information. As a result, they consume significant memory and require extensive training time. In response to these limitations, Szeto et al. \cite{szeto2021hypercon} and Rivoir et al. \cite{rivoir2021long} extended i2i models into video translation frameworks. HyperCon \cite{szeto2021hypercon} ensured temporal consistency by translating temporally interpolated video framewise and aggregating over localized windows. The model in \cite{rivoir2021long} introduced unpaired image translation combined with neural rendering, incorporating global learnable textures and a lighting-invariant view-consistency loss. 

We group these approaches into several categories for better contrasting and make a comparative study between them. We first categorize the SOTA into two broad categories based on the type of input sample: (i) image-to-video (i2v) translation \cite{shen2019facial,fan2019controllable,zhao2020towards}: generating a video from a single or a set of images, (ii) video-to-video (v2v) translation \cite{bansal2018recycle, wang2018video,wei2018video,mallya2020world,wang2019few,bashkirova2018unsupervised,liu2020unsupervised,chen2019mocycle,park2019preserving,wang2022learning,liu2021unsupervised,tulyakov2018mocogan,li2021i2v,elhefnawy2023polygon,szeto2021hypercon,rivoir2021long,chu2020learning,xu2020ofgan,prajwal2020lip,magnusson2021invertable} : generating a video from an input video. Increasing attention has been paid to the video-to-video (v2v) translation framework these days, because it can be used for a wider range of activities than i2v, including translating videos to their appropriate pixel-level semantic segmentation masks, and converting infrared to visible videos, etc. Depending on the mapping between the input and the output samples, we further categorize v2v translation into two types: (a) paired v2v translation \cite{wang2018video, wei2018video,mallya2020world,wang2019few} : data contains one-to-one mapping between input and output video frames, and (b) unpaired v2v translation \cite{bansal2018recycle,bashkirova2018unsupervised,liu2020unsupervised,chen2019mocycle,park2019preserving,wang2022learning,liu2021unsupervised,tulyakov2018mocogan,li2021i2v,elhefnawy2023polygon,szeto2021hypercon,rivoir2021long,chu2020learning,xu2020ofgan,prajwal2020lip,magnusson2021invertable}: the objective is to determine the mapping between the source and the target domains, without knowing the frame-level mapping. Obtaining paired datasets is particularly difficult because it is time consuming, requires subject expertise, and is occasionally impossible in some circumstances. Unpaired v2v translation with the goal of connecting two domains has been offered as a solution to these challenges. As a result, unpaired v2v translation has been receiving a lot of interest from researchers nowadays. Its objective is to determine the mapping between a source and a target domain, without knowing the frame-level mapping. Finding the mapping between two domains along with maintaining the temporal consistency can be challenging. Several approaches have been reported in the literature, including 3DGAN-based approaches such as \cite{bashkirova2018unsupervised}, approaches that explicitly introduce temporal constraints \cite{bansal2018recycle}, those that use optical flow between successive frames \cite{chen2019mocycle,park2019preserving,wang2022learning,chu2020learning,xu2020ofgan}, RNN-based models \cite{liu2021unsupervised,tulyakov2018mocogan}, and approaches that extend i2i into video synthesis \cite{szeto2021hypercon,rivoir2021long}. To improve model training, a variety of objective functions have also been proposed. This survey will include a discussion on the strengths and weaknesses of these methods, explored along with a thorough comparative analysis. The survey will also discuss future research directions.

\section{Image-to-video translation}
\begin{figure}[ht]
    \centering
    \includegraphics[width=0.6\textwidth]{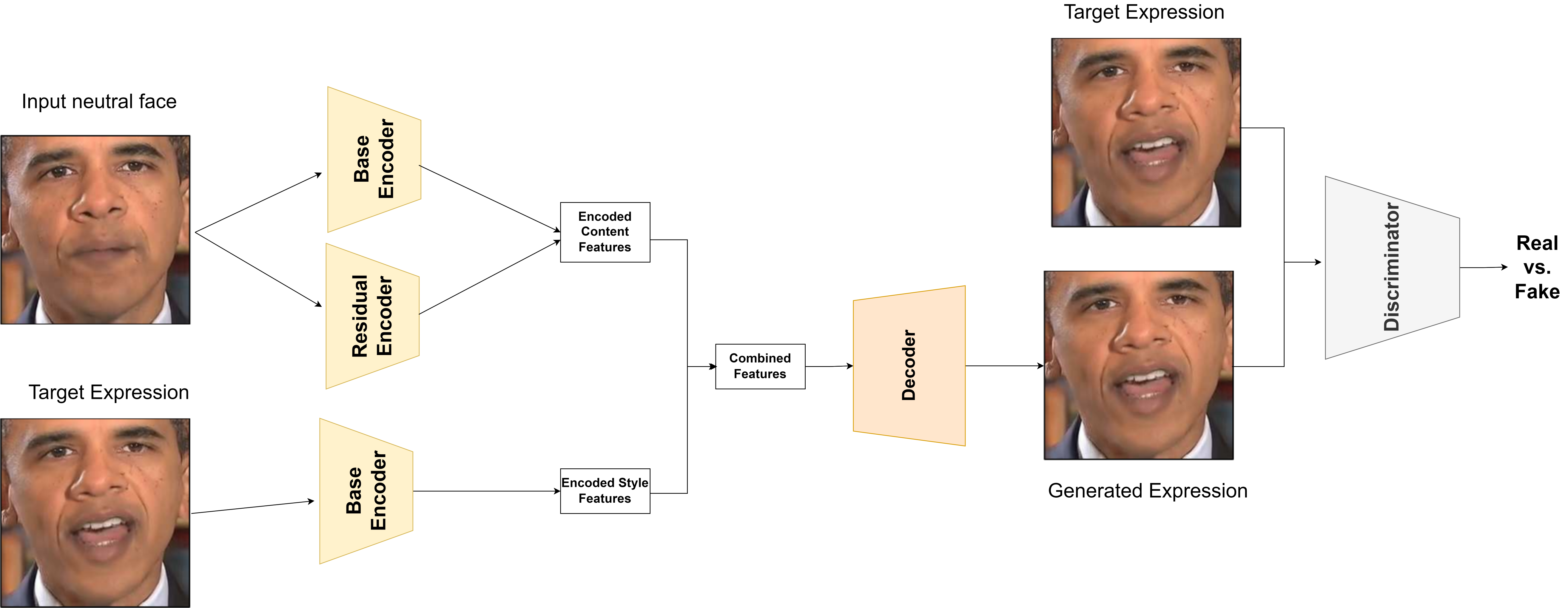}
    \caption{Architecture of AffineGAN. Here, the base and residual encoders extract the content features from the input image, while the second base encoder extracts the style features from the target image. The style and content features are passed to the decoder to generate the target expression video. A discriminator plays role in determining whether the generated expression is real or fake \cite{shen2019facial}.}
    \label{fig:AffineGAN for image-to-video synthesis}
\end{figure}

The realm of visual transformation has witnessed a remarkable evolution with the emergence of image-to-video (i2v) translation. This innovative concept enables the seamless transition of a static image into a dynamic video sequence, thereby expanding the horizon of conventional image manipulation. What sets i2v apart is that the neural networks powering these transformations do not rely on memorization of the training dataset. Instead, they acquire the capacity to generate entirely new visual content, a phenomenon akin to imaginative creation. This intriguing process leverages a concealed affine transformation, enabling the network to produce novel scenes and sequences upon deployment. In the landscape of i2v, various endeavors have honed in on different aspects, ranging from altering body language to manipulating facial expressions. Shen et al. \cite{shen2019facial} presents a remarkable contribution in the field of facial expression animation through the introduction of AffineGAN. This novel approach is capable of accurately predicting facial expression videos of varying temporal lengths from a single static image. The procedure takes a neutral expression and transforms it into a variety of expressions, such as happiness or anger.  One of the distinctive features of AffineGAN is that it can automatically detect the intensity of expression without requiring any other information. The model consists of a residual encoder and a basic encoder. A pair of encoders, one basic and one residual, are employed to record facial contours, while another basic encoder focuses specifically on capturing expression changes, such as mouth movement during a smile. Following the feature extraction, these extracted attributes are combined and fed into a decoder. The primary function of this decoder is to synthesize the target expression in video frames. Additionally, a discriminator is utilized to differentiate between the real and the generated video frames, ensuring the authenticity and accuracy of the generated content. Figure \ref{fig:AffineGAN for image-to-video synthesis} shows the AffineGAN architecture. AffineGAN allows users to control the frame rate for depicting expression changes, making it a versatile approach. The only annotation required is the selection of a neutral face frame per training video. Their model does not need complete, ordered training videos of expressions and can work with incomplete and unordered frames too. However, their model is limited to faces only and struggles to generate realistic backgrounds. Fan et al.  \cite{fan2019controllable} proposed a similar approach that understands how to change facial appearance, such as making the person smile or look surprised based on user input. This network takes the input image and the desired expression change as instructions and produces a modified image that reflects the desired expression. Another important aspect of the methodology is predicting where key points or landmarks are on the face, such as the eyes, nose, and mouth. They used a separate neural network to predict these landmarks on the original and the modified face images. It improves on standard adversarial training by using separate global and local discriminators. However, this approach is limited to faces and simple backgrounds. While prior methods largly focused on facial expressions, Zhao et al. \cite{zhao2020towards} proposed a two-stage model that can perform both facial expression retargeting and human pose forecasting. A neural network predicts a sequence of structural conditions, such as facial expressions or body poses, that guide how the object should move over time. A second network then uses those structural conditions to generate a coarse video where the object moves according to the sequence. Finally, a third network refines the coarse video to make the motions look more smooth and temporally coherent. One challenge is that the adversarial loss can lead motion refining network to an suboptimal identity mapping solution between the input and the output videos. To address this, the authors propose a ranking loss that optimizes over distance comparisons between the refined video, the input video, and the target video. Their model combines the benefits of conditioning on spatial structures and using 3D CNNs for temporal smoothing.

\section{Video-to-video translation}
Drawing inspiration from the prior advancements, a range of innovative video-to-video translation (v2v) methods have emerged. These methods have been developed to address challenges in diverse applications, including video restoration, video colorization, and video segmentation, etc. The motivation behind these methods lies in their ability to transform one video into another while maintaining the underlying content structure, allowing for the enhancement, manipulation, or modification of video content.
In the context of these v2v approaches, there is a noticeable evolution from earlier methodologies to more recent ones. The early methods typically operated under the constraint of needing paired video frames that correspond between the input and output videos. This requirement of frame-by-frame correspondence was often met by manually or algorithmically associating each frame in the input video with a corresponding frame in the output video. 
However, as the field progressed, later-generation of v2v methods have aimed to alleviate the dependency on such paired video frames. These advanced approaches have sought to break free from the rigid requirement of explicit frame correspondences between the input and the output videos. Instead , they have harnessed more flexible strategies, such as leveraging unpaired data or exploring the relationship between different video frames without relying on a strict one-to-one frame correspondence.
To further categorize these v2v techniques, a useful distinction can be made based on the nature of input and output data. Specifically, the existence of labeled input and output frames forms the foundation for this categorization. In this scheme, v2v methods can be divided into two categories, paired and unpaired.

\subsection{Paired v2v}

\begin{figure}[ht]
    \centering
    \includegraphics[width=0.8\textwidth]{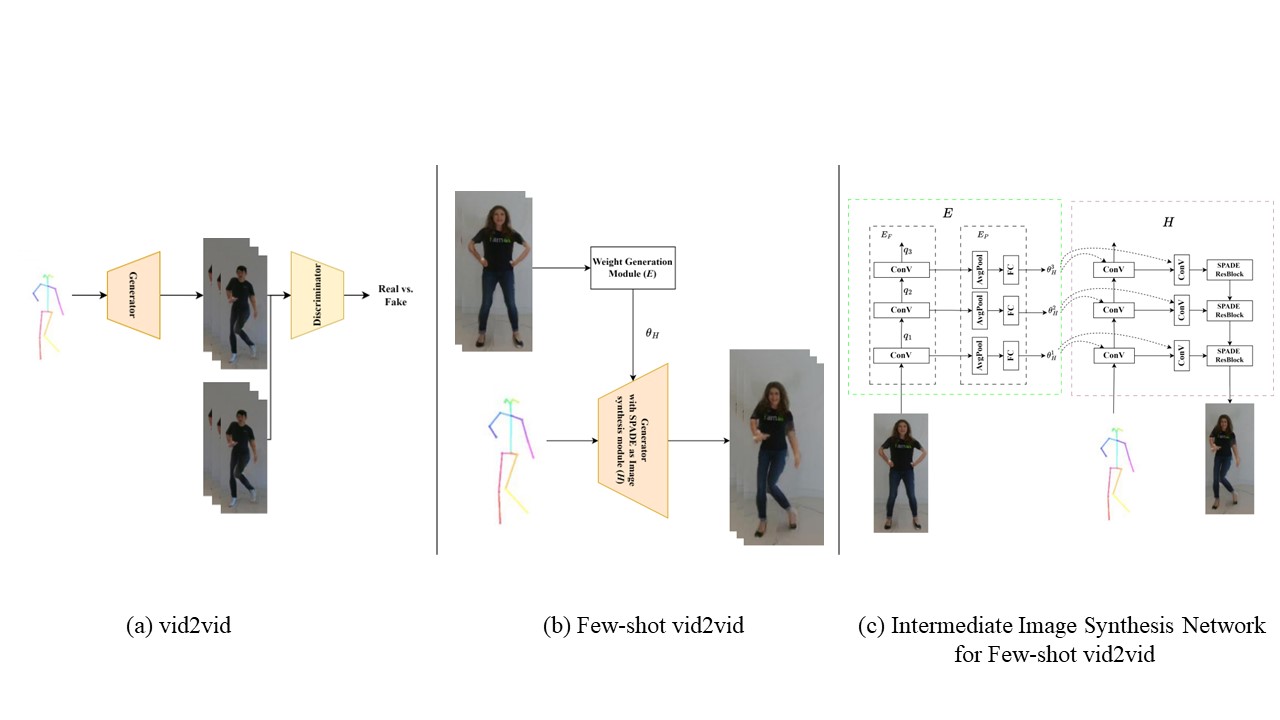}
    \caption{Paired v2v for video synthesis (a) Architecture of vid2vid framework. A conditional generator is used as image synthesis unit ($H$) \cite{wang2018video}, (b) Architecture of the few-shot vid2vid framework. A network weight generation module $E$ is used to generate weight from example image ($e$). Modified SPADE was used for $H$ \cite{wang2019few}, (c) Module $E$ consists of sub-network $E_F$ for extracting features $q$ from $e$, and sub-network $E_P$ to generate wegith $\theta_H$ from $q$ for $H$ \cite{wang2019few}.}  
    \label{fig:Paired v2v for video synthesis}
\end{figure}

\begin{figure}[ht]
    \centering
    \includegraphics[width=0.8\textwidth]{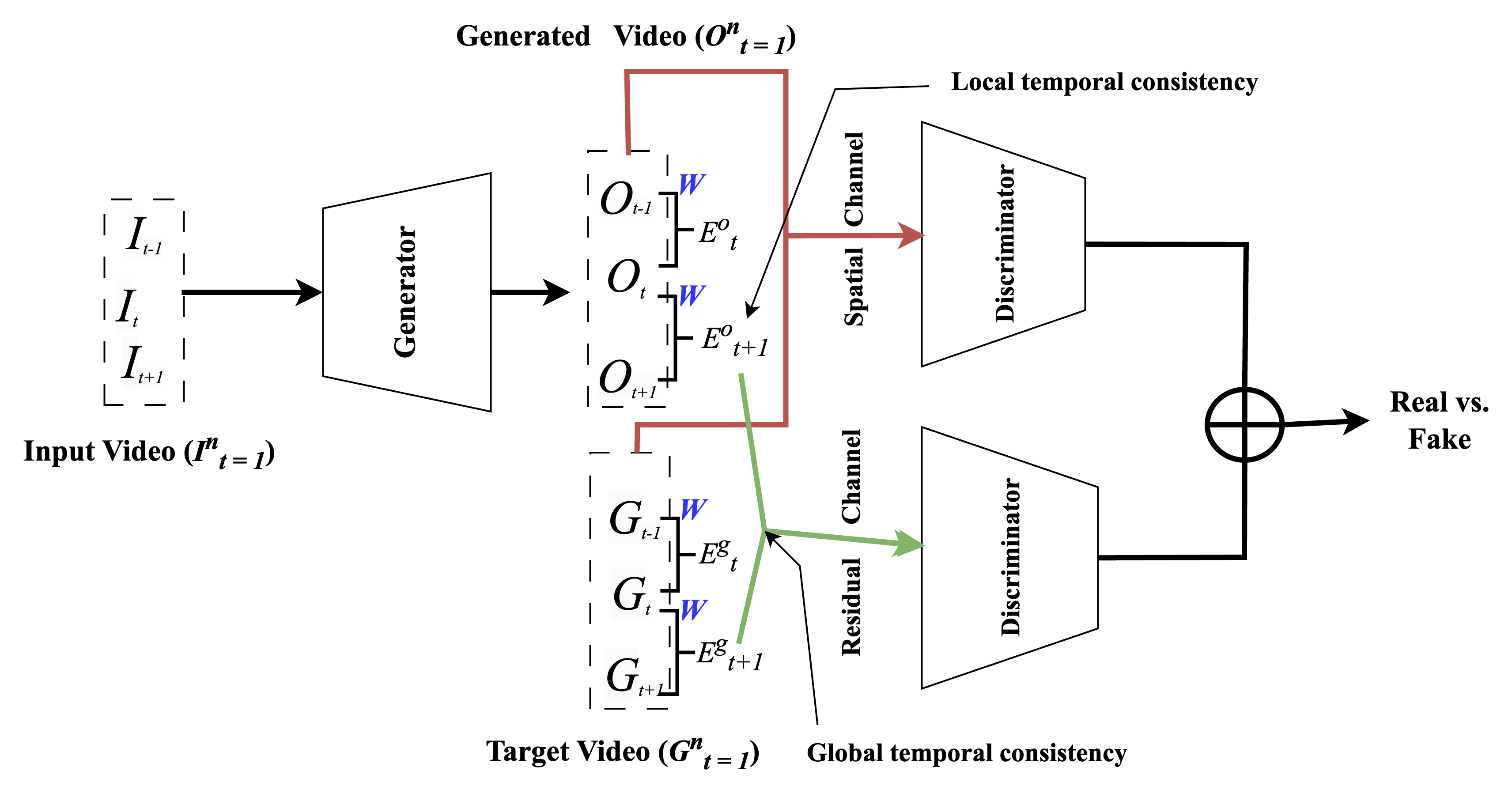}
    \caption{This figure presents the architecture of vid2vid with global temporal consistency. The input video sequence is denoted as \(I = [\ldots, I_{t-1}, I_t, I_{t+1}, \ldots]\), and the corresponding ground-truth is represented as \(G = [\ldots, G_{t-1}, G_t, G_{t+1}, \ldots]\). The output from the generator is given by \(O = [\ldots, O_{t-1}, O_t, O_{t+1}, \ldots]\). The frame \(O_{t-1}\) is warped using optical flow (\(W\)) to produce \(O'_{t}\), aligning it with \(O_t\). Residual errors are computed for both the ground-truth (\(E^g_t, E^g_{t+1}\)) and the generated output (\(E^o_t, E^o_{t+1}\)). A two-channel discriminator is employed, with one channel distinguishing between the generated and ground-truth output and the other discerning the residual errors between the ground-truth and predicted videos \cite{wei2018video}.}
    
    \label{fig:wei}
\end{figure}

These methods tackle video-to-video translation by treating it as a problem of matching distributions. The main goal is to train a model to make the output videos look as natural and similar as possible to real videos from the input ones. The vid2vid technique \cite{wang2018video}, for instance, employs a conditional generative adversarial model and introduces a novel spatio-temporal objective  function. The primary objective of this method is to effectively manage paired input-output videos. It accomplishes this task by employing a sequential generator, which produces video frames in a step-by-step fashion. In this process, a feedforward network (F), plays a crucial role, which predicts the likelihood of the current output frame based on the two preceding input and generated frames. Figure \ref{fig:Paired v2v for video synthesis}(a) illustrates the architecture of the vid2vid model. This framework operates by processing a sequence of video frames through a sequential generator. The primary role of the generator is to synthesize video sequences that emulate the input frames. Concurrently, a discriminator is employed to evaluate these generated videos. It functions by discerning the disparities between the videos synthesized by the generator and the ground-truth video sequences. This method faces challenges in maintaining consistent motion dynamics and smooth transitions over longer sequences, resulting in potential jerkiness and unrealistic transitions between frames. Additionally, it might exhibit difficulties in preserving natural motion characteristics, leading to generated videos that appear more like computer-generated animations than realistic recordings. Wei et al. \cite{wei2018video} improve upon these limitations by introducing the concept of global temporal consistency. This improvement addresses the aforementioned limitations by ensuring smoother motion transitions, improving frame-to-frame coherence, and providing a more natural representation of real-world motion and scene composition. This method employs residual errors to achieve this goal, calculating these errors separately for both source and generated videos through a comparison of warped and aligned frames. The discriminator architecture consists of two separate channels: one encodes information related to ground-truth and generated frames, while the other channel encodes information about residual errors resulting from comparisons between predicted and actual outputs. The outputs from these two channels are then merged to yield the discriminator's final output. The proposed architecture is illustrated in Figure \ref{fig:wei}.

Although these methods ensure short-term temporal consistency, they falter in maintaining long-term consistency due to their reliance on a sequential frame generation process. Moreover, they may not fully encapsulate the complex 3D dynamics of the world, possibly leading to inaccuracies in motion and scene depiction. To address long-term temporal consistency in video generation, Mallya et al. \cite{mallya2020world} introduced a novel framework for video-to-video synthesis. This method utilizes all previously generated frames during rendering by synthesizing the current frame conditioned on all preceding frames using a `guidance image'. This image is a physically-grounded estimate of the subsequent frame's appearance, based on the rendered scene's 3D structure. The guidance images are derived by first reconstructing a 3D point cloud of the scene's static elements using structure from motion (SfM) \cite{longuet1981computer,tomasi1992shape}. As the video progresses frame by frame, the output frames are back-projected onto this 3D point cloud to texture it. The textured point cloud is then projected to create a guidance image for the upcoming frame, considering the estimated camera viewpoint. This is achieved using a multi-SPADE \cite{park2019semantic} module. Their innovative design ensures consistency across the entire video sequence, merging concepts from conditional image synthesis models \cite{park2019semantic} and scene flow \cite{vedula1999three}, thus enhancing the long-term coherence and accuracy of motion representation in extended sequences.

The preceding approaches encountered limitations in their ability to synthesize videos with high fidelity for unseen domains or subjects due to their reliance on paired training data. To address these limitations, Wang et al. \cite{wang2019few} introduce a novel few-shot framework that facilitates the synthesis of videos for new and unseen subjects or scenes. This is achieved by utilizing a small number of examples from the target domain during testing. This framework requires two inputs for the generator network (Figure \ref{fig:Paired v2v for video synthesis}(b)), a standard semantic video input, represented as \(S = \{s_1, s_2, \ldots, s_n\}\), where $s_t$ (for \(1 \leq t \leq n\)) denotes the $t$-th frame in a sequence of $n$ total frames, as used in the conventional vid2vid approach \cite{wang2018video}, and a set of example images (e.g., $e_1,e_2,....,e_i$) from the target domain, provided during the testing phase. Wang et al. \cite{wang2019few} replace the conditional image generator, used in vid2vid models, with a modified SPADE (Spatially-Adaptive Normalization) generator \cite{park2019semantic} to build their intermediate image synthesis network ($H$). The core idea of SPADE architecture is to preserve the essential structure of the content (e.g., the skeletal mask in Figure \ref{fig:Paired v2v for video synthesis}(b)) while allowing flexibility in the style aspects (e.g., example images in Figure \ref{fig:Paired v2v for video synthesis}(b)) of the output. The key functionality of SPADE is to keep the content information consistent, while the style can be varied by altering the input noise. This enables the generation of diverse outputs with the same underlying structure but different stylistic appearances \cite{park2019semantic}. The image synthesis network \(H\), consisting of \(L\) layers, is paralleled by a weight generation network \(E\) structured similarly with \(L\) layers. $E$ incorporates a multi-layer perceptron ($E_P$) and an example feature extractor ($E_F$) consisting of several convolutional layers. An appearance representation ($q$) is extracted from an example image ($e_t$) using the network $E_F$ and subsequently fed into $E_P$ to get the weights $\theta_H$ for $H$. When there are multiple example images, the weighted average of extracted features from them is used to integrate the obtained features. $\theta_H$ has three sets of weights: $\theta_S$, \(\theta_{\gamma}\), and \(\theta_{\beta}\). $\theta_S$ is subsequently employed to convolve each frame ($s_t$) derived from the input semantic video ($S$)
and generate a normalized feature ($P_S$). \(\theta_{\gamma}\) and \(\theta_{\beta}\) convolve $P_S$ to generate scale $\gamma$ and bias $\beta$, respectively. The difference between the proposed SPADE module and the original SPADE module is that the original SPADE module is designed to modulate the normalization parameters spatially across \(s_t\), which is in this method enhanced by the weight generation module to adaptively convolve \(s_t\), and generate scale \(\gamma\) and bias \(\beta\) parameters for affine transformations in each layer of the network. The generator denormalizes the normalized feature ($P_S$) at each label to a denormalized feature ($P_H$) by ($P_H = P_S * \gamma + \beta$). The denormalized features maintain the semantic information from the semantic mask across the whole network and generate the output frame with content information from the semantic mask and style information from the provided example images. This innovative approach enables the model to adapt to new scenes and subjects at test time, a capability not present in earlier methods, which were confined to synthesizing videos similar to their training data. This method is applicable to synthesizing dynamic human motions and actions, which enables the creation of lifelike human animations and actions that can be integrated into various digital media formats. Additionally, the method is adept at transforming semantic segmentation masks into photo-realistic videos (\textbf{label-to-video}), and videos to corresponding semantic segmentation masks (\textbf{video-to-label}) within unexplored domains, enhancing the versatility of video generation in diverse scenarios.

\subsection{Unpaired v2v}
Paired v2v methods necessitate the annotation of each individual frame within a video, an endeavor that becomes increasingly arduous given the substantial number of frames typically found in standard videos. For instance, a video with a frame rate of 30 frames per second (30fps) accumulates a significant number of frames over its duration. This annotation process becomes not only labor-intensive but also time-consuming, making it a challenging task for paired v2v methods. This challenge is particularly pronounced in scenarios where videos are of extended length or when large-scale datasets are involved. To address this limitation, a range of innovative approaches have emerged, shifting the focus towards unpaired v2v methods. These methods aim to bypass the strict requirement of frame-by-frame annotation by devising strategies that utilize unpaired data more efficiently. Unpaired methods embrace the inherent relationships between different video domains and leverage this intrinsic information to enable translation without explicit one-to-one correspondence between input and output frames. To further clarify the landscape, these unpaired v2v methods can be categorized into five distinct subcategories. This categorization provides a structured framework for understanding and analyzing the diverse strategies employed in unpaired methods, helping to identify their strengths, limitations, and potential for advancing the field of video-to-video synthesis.

\subsubsection{3D GAN-based approach}
\begin{figure}[ht]
    \centering
    \includegraphics[width=0.6\textwidth]{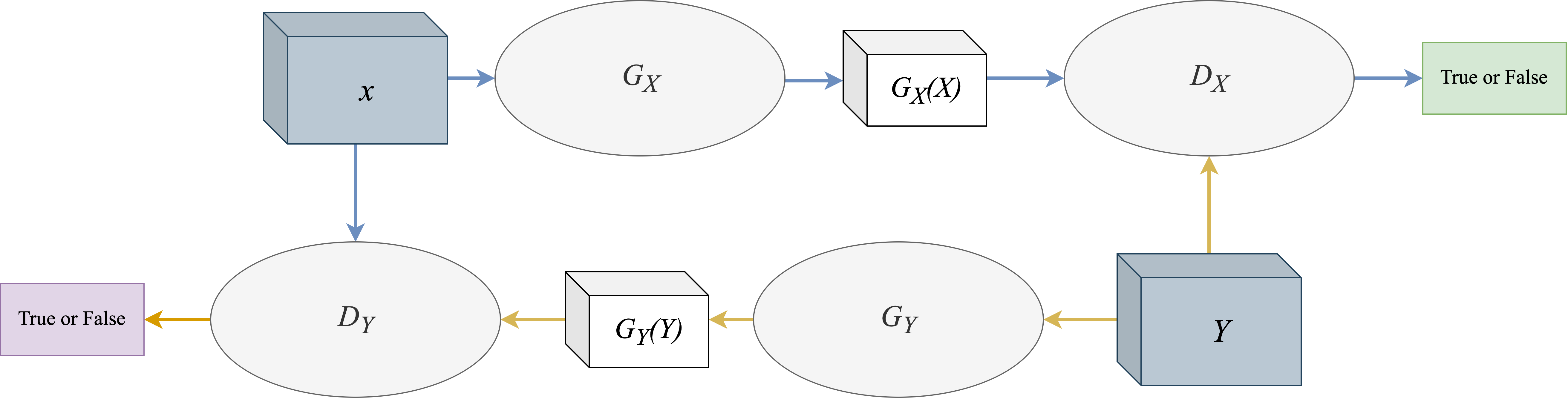}
    \caption{3D GAN for video synthesis. The model features two generator networks, $G_X$ and $G_Y$, designed to convert volumetric images ($X$ and $Y$) from one domain to another. Additionally, there are two discriminator networks, $D_X$ and $D_Y$, tasked with differentiating between real and synthetic videos. A key aspect of the model is its cycle consistency, ensuring that an image translated to another domain and then back again by $G_Y$($G_X$($X$)) remains identical to the original input \cite{bashkirova2018unsupervised}.} 
    \label{fig:3D GAN for video synthesis}
\end{figure}

A 3D conditional generative adversarial network (GAN) is proposed by \cite{bashkirova2018unsupervised} that treats both input and output data as three-dimensional tensors. By structuring the data in this 3D tensor format, the network is capable of effectively processing image sequences while accounting for the temporal dimension. The design of the architecture allows it to efficiently handle sequences of frames, by considering them as 3D tensors with dimensions \textit{d}  × \textit{h} × \textit{w}. Here, \textit{d} represents the temporal depth or length of the sequence, while \textit{h} and \textit{w} denote the height and width of individual images within the sequence, respectively. The significance of this approach lies in its ability to exploit the inherent temporal relationships between frames. By treating the video data as a cohesive 3D tensor, the network can capture both the spatial information within individual frames and the temporal dynamics spanning the sequence. Figure \ref{fig:3D GAN for video synthesis} presents the architecture of a 3D Generative Adversarial Network (GAN) used for video synthesis. The model encompasses two generator networks, \(G_X\) and \(G_Y\), which are designed for the transformation of volumetric images between two domains, \(X\) and \(Y\). The generators are complemented by two discriminator networks, \(D_X\) and \(D_Y\). Their role is to discern between real and synthetically generated videos. A pivotal feature of this model is cycle consistency. This aspect ensures that a video, once translated to a different domain by \(G_Y\), can be reverted by \(G_X(G_Y(X))\), thus retaining its originality. This setup helps in effectively converting videos from one style to another while ensuring that the generated videos are as close to real as possible.
\subsubsection{Temporal constraint-based approaches}
The 3D GAN-based approach for video-to-video translation has limitations due to its narrow focus on the processing of image sequences for video-to-video translation. While the network’s utilization of a 3D tensor format is innovative and effective for incorporating temporal dimensions, it does not inherently address the challenges of aligning and translating the content between different domains, which is the primary goal of video-to-video translation tasks. Also, since the 3D tensor approach attempts to directly encode temporal information within the data representation, it can be burdened with complexity and scalability issues. Temporal constraint-based methods, on the other hand, focus on effectively capturing temporal relationships using more streamlined and adaptable approaches, leading to potentially more manageable models with improved training efficiency and performance.

\begin{figure}[ht]
    \centering
    \includegraphics[width=0.6\textwidth]{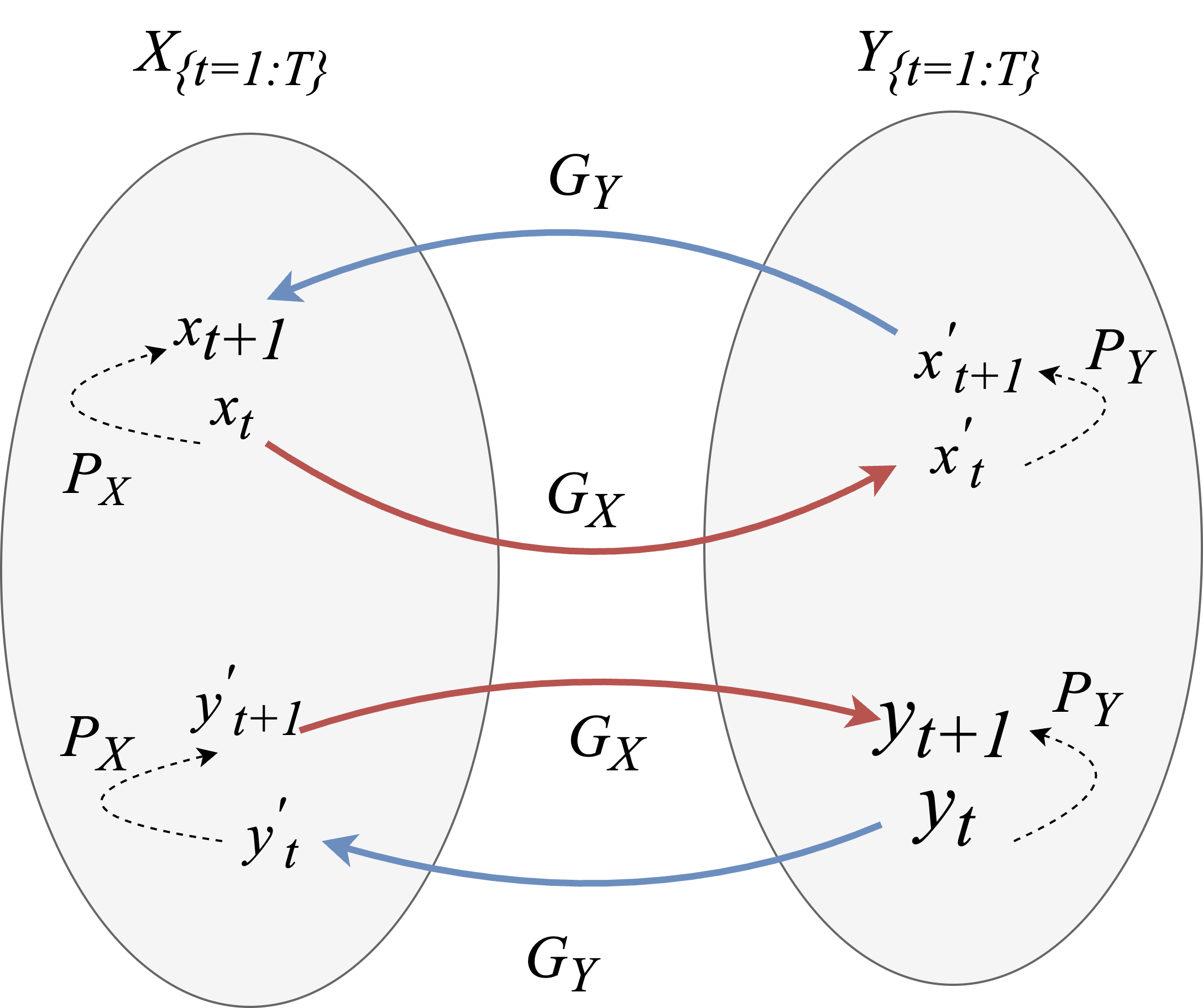}
    \caption{Illustration of the Recycle-GAN approach for video synthesis. The framework operates on two distinct but sequentially linked data streams, represented as \(x = [x_1, x_2, \ldots, x_t]\) and \(y = [y_1, y_2, \ldots, y_s]\). Within this setup, \(G_X\) and \(G_Y\) function as generators tasked with creating synthetic video frames. Complementing these, \(P_X\) and \(P_Y\) serve as temporal predictors, designed to recognize and utilize the temporal relationships inherent in these ordered sequences \cite{bansal2018recycle}.}
    
    \label{fig:Temporal predictor for video synthesis}
\end{figure}

Recycle-GAN \cite{bansal2018recycle} introduces a methodology for video retargeting and unpaired v2v translation using spatio-temporal constraints along with conditional generative adversarial networks (cGANs) \cite{goodfellow2020generative}. The authors show that using both spatial and temporal constraints is more advantageous than employing spatial constraints alone. They introduce a recurrent temporal predictor to capture and exploit the temporal dependencies present in ordered sequences of data. The predictor uses past frames within a sequence to predict the subsequent frame in the sequence. The predictor is trained using recycle-loss function that measures the difference between the predicted frame and the actual frame. This loss encourages the predictor to capture meaningful temporal relationships and patterns in the data. The Recycle-GAN maintains a cycle across domains and time, by minimizing the L1 loss between the generated frame and the real frame. However, this approach may not fully capture the modification tendencies and trends present in the target frames since it only improves the performance of future frame generation based on predicted and ground-truth frames without considering the pattern of frame change over time. Therefore, it may struggle to accurately preserve the nuanced patterns and changes between adjacent frames. Figure \ref{fig:Temporal predictor for video synthesis} depicts the Recycle-GAN framework \cite{bansal2018recycle}, a method focused on video synthesis through temporal constraints. This framework processes two sequential data streams, denoted as \( x = [x_1, x_2, \ldots, x_t] \) and \( y = [y_1, y_2, \ldots, y_s] \). In this architecture, \( G_X \) and \( G_Y \) are the generators responsible for producing synthetic video frames. These generators are complemented by \( P_X \) and \( P_Y \), which act as temporal predictors. The role of these predictors is to identify and leverage the temporal correlations present within these sequential data streams.The interplay between the generators and the temporal predictors is a key aspect of Recycle-GAN. While the generators focus on the visual fidelity of the frames, the temporal predictors work to maintain a coherent temporal flow. This dual approach ensures that the synthesized videos exhibit a high degree of realism, both in terms of individual frame quality and the overall sequence continuity. Liu et al. propose a new method \cite{liu2020unsupervised} based on the Recycle-GAN framework by introducing a novel loss called the ``tendency-invariant loss''. This loss is specifically designed to address the limitation of Recycle-GAN, which is the inadequate preservation of modification trends between frames. The tendency-invariant loss takes into account not only the modifications between adjacent frames but also the inherent modification tendencies observed in the generated frames. This loss aims to capture and ensure the consistency of the way frames change over time. More details on this loss are provided in section \ref{sec:Objective Functions}. By incorporating this loss, the new method ensures that the modifications present in the target frames are accurately preserved and propagated to the generated frames.

\subsubsection{Optical flow-based methods}
\begin{figure}[ht]
    \centering
    \includegraphics[width=0.6\textwidth]{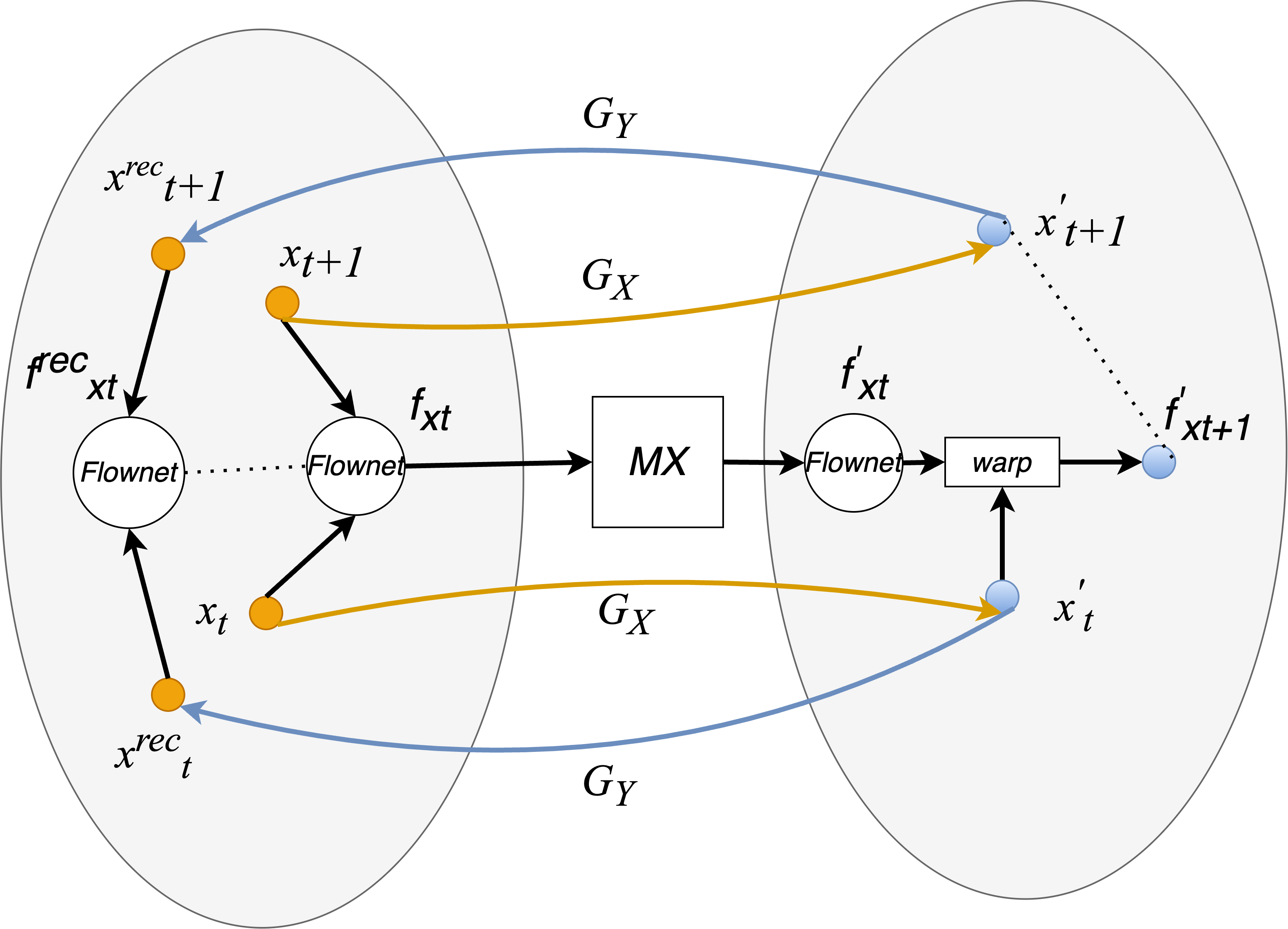}

    \caption{This figure demonstrates an optical flow-based method for processing two distinct data sets, denoted as \(x = [x_1, x_2, \ldots, x_t]\) and \(y = [y_1, y_2, \ldots, y_s]\). For simplicity, the figure only includes generation of domain $y$ from domain $x$.The method employs an optical flow technique (referred to as `warp') to analyze and model motion between successive frames. The warp function is instrumental in capturing the motion flow between two consecutive frames and in predicting the subsequent frame. This function plays a pivotal role in transferring motion information across the two data domains, \(X\) and \(Y\), thereby significantly improving the consistency and continuity of motion in video synthesis \cite{chen2019mocycle,park2019preserving}.}
    \label{fig:Optical flow for video synthesis}
\end{figure}

The challenge of precisely maintaining small variations across successive video frames is one of the fundamental problems with the temporal constraint-based method. These techniques are likely to have trouble capturing finely detailed patterns and alterations in the video frames, producing less realistic outcomes. Moreover, they could have trouble scaling when working with longer video sequences or huge datasets, decreasing their effectiveness. This section aims to address these limitations by introducing optical flow-based methods that focus on improving the accuracy and consistency of motion representation, thus enhancing the overall quality of video translation. Optical flow is a technique used in computer vision to estimate the motion of objects between consecutive frames of a video. It provides information about the direction and magnitude of pixel movements between frames, which can be used to understand how objects are moving within the video. Flownet \cite{dosovitskiy2015flownet} and Flownet 2.0 \cite{ilg2017flownet} are two deep learning architectures specifically designed for optical flow estimation. They extract meaningful features from input frames and learn to predict the direction and magnitude of pixel movement between them. They accomplish this by matching the features extracted from both frames, allowing it to estimate optical flow maps that capture motion dynamics. Figure \ref{fig:Optical flow for video synthesis} illustrates the architecture of an optical flow-based method for video synthesis. The figure represents the processing of two separate data sets, symbolized as \(x = [x_1, x_2, \ldots, x_t]\) and \(y = [y_1, y_2, \ldots, y_s]\). Central to this method is the implementation of an optical flow technique, often referred to as `warp'. This technique is crucial for analyzing and modeling motion between successive frames. The warp function is particularly important in tracking the motion flow between two consecutive frames and is instrumental in predicting the next frame in the sequence. Its significance lies in its ability to transfer motion information effectively between the two data domains, \(X\) and \(Y\). This transfer enhances the overall consistency and continuity of motion, thereby markedly improving the quality of motion representation in synthesized video sequences.
Mocycle-GAN \cite{chen2019mocycle} integrates motion estimation and specialized constraints to enhance video translation quality across unpaired domains. The motion translator (MT) is a unique addition in Mocycle-GAN, which is responsible for translating motion information between the source and the target domains. Motion information is captured through optical flow computed by FlowNet \cite{dosovitskiy2015flownet}, which illustrates how pixels move between frames. Mocycle-GAN leverages three training constraints to ensure effective translation: the adversarial constraint for realistic frame appearance, frame and motion cycle consistency constraints for maintaining appearance and motion dynamics, and the motion translation constraint utilizing MT to transfer motion between domains and enforce temporal continuity in synthetic frames. Park et al.  \cite{park2019preserving} address the limitation of temporal constraint-based methods by utilizing optical flow to incorporate both spatial and temporal information effectively. Instead of using a future frame predictor, the new approach warps the previously generated frame using optical flow and combines it smoothly with the currently translated frame. The proposed method introduces a recurrent generator comprised of three essential sub-modules: an image generator, a flow estimator, and a fusion block. The function of the recurrent generator is auto-regressive, and the entire video output is generated by iteratively applying it. In each step, the image generator generates an image, and a flow estimator calculates the optical flow between consecutive frames. For regions of pixels affected by occlusions or new scene elements where optical flow is unreliable, a fusion block generates a soft mask that adaptively combines warped and generated pixels into one frame. However, the mentioned approaches struggle with a significant limitation in their reliance on optical flow estimators, which are prone to producing inaccurate motion estimations due to model capacity constraints and domain gap issues. To address this limitation, \cite{wang2022learning} introduces a solution involving synthetic optical flow, which circumvents the inaccuracies associated with motion estimation. This synthetic optical flow is used to represent flawless motion across domains, ensuring that the regularization process enforces consistent motion without the risk of incorrect motion estimation. This method involves the synthesis of random optical flow for static frames in the source domain, allowing for the simulation of temporal motion and subsequent frame synthesis through warping. The same synthetic optical flow is used to simulate the future frame for the target domain counterpart translated from the source frame. Spatio-temporal consistency regularization is then applied between each  pair of future frames, maintaining motion consistency across domains.

\subsubsection{RNN-based methods}
\begin{figure}[ht]
    \centering
    \includegraphics[width=0.6\textwidth]{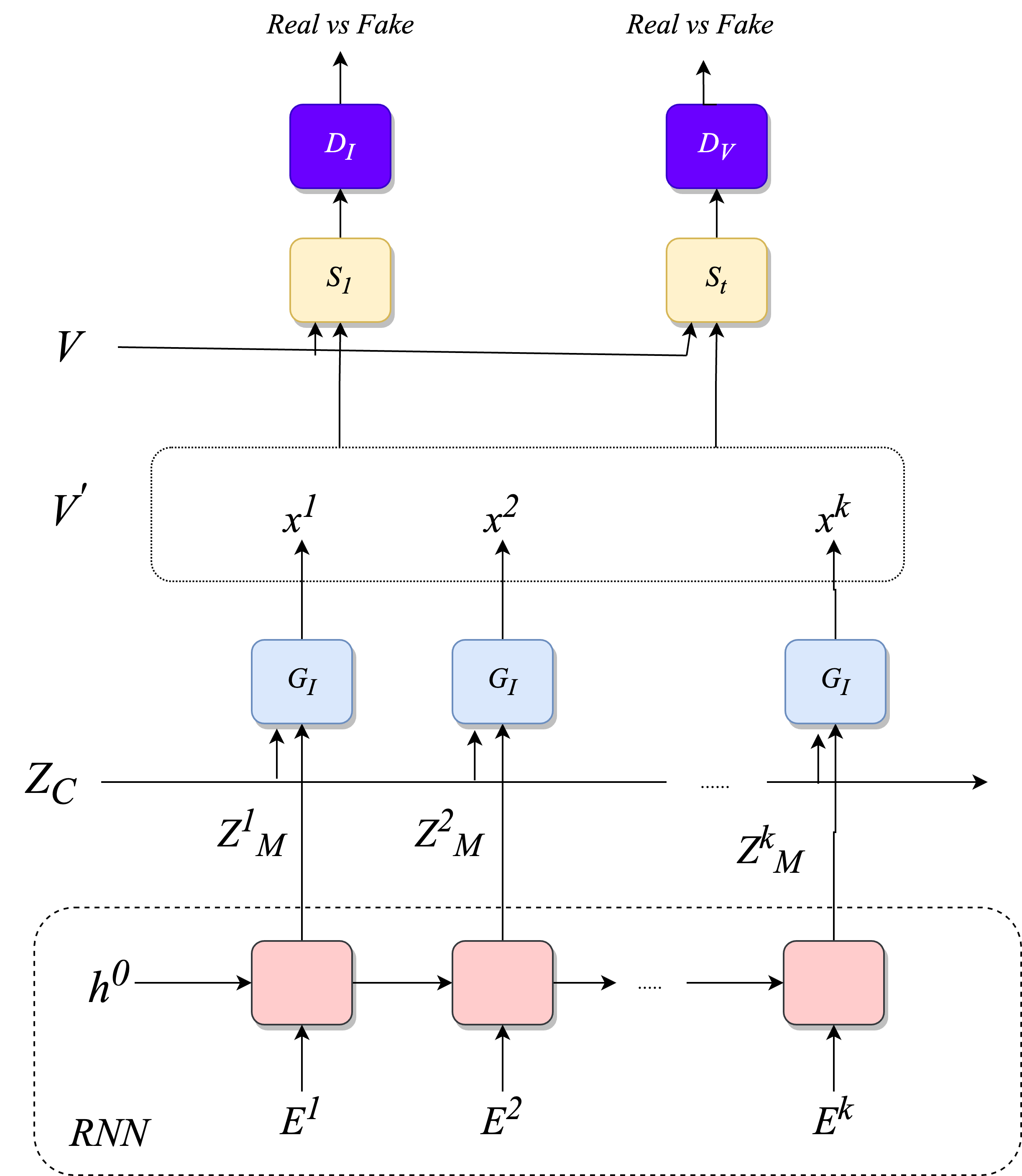}

    \caption{This figure demonstrates the MoCoGAN framework for video generation. The process begins with the selection of a constant content vector \(Z_C\), which remains unchanged for the entire video sequence. A recurrent neural network (RNN) generates a sequence of random variables \([E_1, \ldots, E_T]\), which are then transformed into a series of motion codes \([Z^1_C, \ldots, Z^T_C]\). These motion codes, combined with the static content vector \(Z_C\), are fed into a generator \(G_I\) to produce individual video frames \(X_k\). The framework incorporates two discriminators, \(D_I\) for images and \(D_V\) for videos. These discriminators are trained to differentiate between real and synthetically generated images and videos from the training dataset \(V\) and the generated set \(X'\), respectively \cite{tulyakov2018mocogan}.}
    \label{fig:RNN for video synthesis}
\end{figure}

Previous techniques for translating videos to videos have certain drawbacks.  Image-to-image methods, such as Cycle-GAN, produce flicker artifacts because they ignore the coherence and consistency of the video. Small 3D convolutions used by 3DGAN are unable to capture long-range dependencies, leading to uneven translations throughout a video.  Recycle-GAN cannot directly incorporate inter-frame context because it depends on a different prediction module. As a result, the generated video content can abruptly shift between modes that do not match, such as suddenly shifting from a sunny scene to a snowy scene in consecutive frames. This inconsistent video content over time is referred to as ``mode collapse''.  UVIT \cite{liu2021unsupervised} aims to overcome these limitations by introducing Unsupervised Multimodal Video-to-Video Translation via self-supervised learning. UVIT uses an encoder-RNN-decoder architecture to decompose videos into content and style representations. The content encoder and decoder aim to preserve semantic information, while the style encoder adds domain-specific characteristics. It uses TrajGRU \cite{shi2017deep} to propagate content information across frames bidirectionally. It trains the model to reconstruct frames from the surrounding context in a self-supervised manner. This provides strong temporal consistency. A similar approach is adopted by MoCoGAN \cite{tulyakov2018mocogan}. MoCoGAN also disentangles content and motion (style) factors. Figure \ref{fig:RNN for video synthesis} presents the MoCoGAN framework. The framework initiates with the selection of a constant content vector \(Z_C\), which remains consistent throughout the entire video sequence. A Recurrent Neural Network (RNN) is utilized to generate a sequence of random variables \([E_1, \ldots, E_T]\), which are subsequently combined with a series of motion codes \([Z^1_C, \ldots, Z^T_C]\). They are passed into a generator \(G_I\) to produce individual video frames \(X_k\). The architecture also includes two discriminators, \(D_I\) for images and \(D_V\) for videos. These discriminators are specifically trained to distinguish between real and artificially generated images and videos, comparing the training dataset \(V\) and the generated set \(V'\). UVIT seems better suited for semantic-preserving tasks such as segmentation while MoCoGAN can synthesize more natural videos by explicitly modeling motion dynamics. A key difference is that UVIT proposes more complex TrajGRU units and introduces an additional self-supervised video interpolation task to address the challenges of training RNN architecture. This provides pixel-level supervision to help train the RNN. In contrast, MoCoGAN uses a simpler Gated Recurrent Unit (GRU) but disentangles motion and content into separate latent spaces for more explicit control over video generation.

\subsubsection{Extended i2i translation-based approaches}
\begin{figure}[ht]
    \centering
    \includegraphics[width=0.6\textwidth]{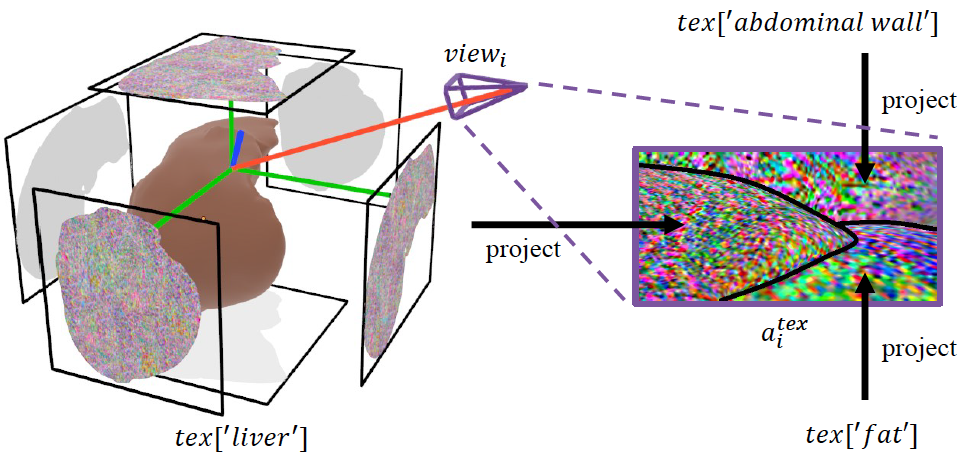}
    \caption{Neural Texture Projection Technique: for each pixel, a ray (illustrated in red) is projected towards the surface of the scene. Utilizing triplanar mapping (depicted in green), points on the surface are correlated with learnable texture planes, incorporating bilinear interpolation within the texture space. This process, being differentiable, facilitates the backpropagation of discrepancies from the image domain to the overarching texture space, allowing for iterative refinement and error correction \cite{rivoir2021long}.}
    \label{fig:viewpoints}
\end{figure}

\begin{figure}[ht]
    \centering
    \includegraphics[width=0.6\textwidth]{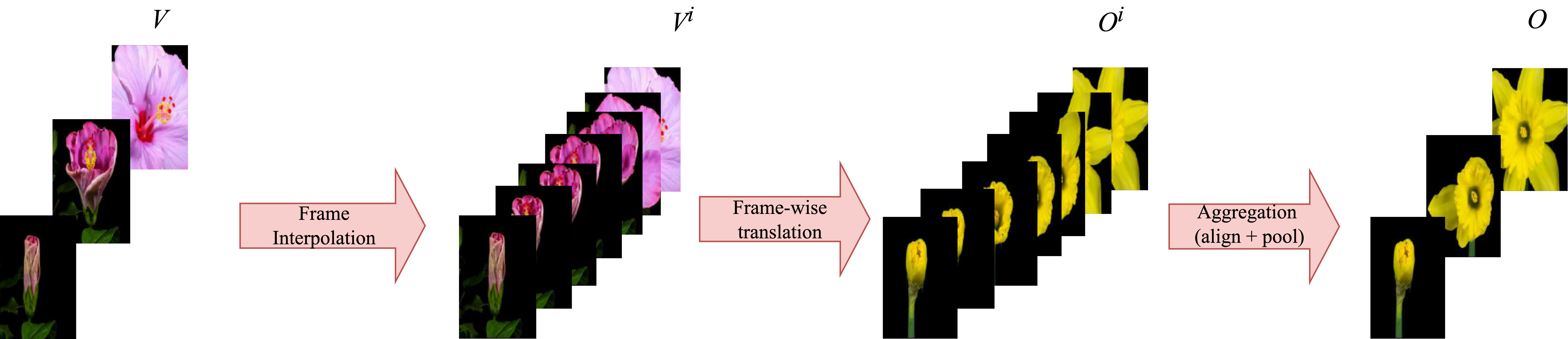}
   \caption{This figure illustrates the Hyperconsistency (HyperCon) technique in a simplified form. The process begins with the enhancement of the input video \(V\) by augmenting it with additional frames using a frame interpolation network, yielding a smoother interpolated video \(V^i\). Each frame of \(V^i\) is then processed individually through an image-to-image translation model. The final step involves a meticulous combination of these frames. This combination is achieved by first aligning the frames using optical flow and then implementing a pixel-wise pooling strategy with a local sliding window. The end result is the output video \(O\), characterized by significantly enhanced temporal consistency \cite{szeto2021hypercon}.}
    \label{fig:i2i for video synthesis}
\end{figure}

In contrast to most previous approaches that introduced novel and intricate architectures for video-to-video (v2v) translation tasks, there exists a notable alternative. The approach highlighted in HyperCon \cite{szeto2021hypercon} takes a different route by leveraging traditional image-to-image (i2i) architectures for v2v translation tasks.  HyperCon follows three main steps. First, it uses frame interpolation to insert additional frames between the original input frames. This creates an expanded interpolated video with more frames. Next, it applies the image translation model independently to each interpolated frame to get a translated but inconsistent video. Finally, it aggregates nearby frames in this translated video using optical flow warping and pooling. This aligns similar content and selects consistent components across frames to output a temporally coherent video. The core idea is that by synthesizing additional interpolated frames and aggregating their translations, HyperCon can pass information between neighboring frames and filter out inconsistent artifacts of the frame-level translation model. Figure \ref{fig:i2i for video synthesis} shows the Hypercon architecture for video synthesis. However, these methods inherently lack suitability for maintaining long-term consistency. Long-range temporal dependencies are not taken into account by HyperCon. Another drawback is that it relies on errors from the frame-wise model (i2i model) it employs, which may sometimes propagate downstream and result in problems with the output. Therefore, HyperCon may propagate errors made by the frame-wise model.
For instance, it can be difficult to maintain smooth transitions when an object disappears and then reenters the frame because the prior frames lack the necessary information regarding the object's appearance. The issue persists in longer recordings, even with multiple frames used to help; therefore, the smoothness of HyperCon is insufficient for prolonged scenarios. Rivoir et al. \cite{rivoir2021long} address the limitaion of previous approaches in generating temporally consistent longer sequences by proposing a novel approach to generate long-term temporally consistent and photo-realistic video translations. The authors use neural rendering in conjunction with unpaired image translation. The key idea is to use a learnable global texture representation to hold appearance information such as material properties, making it possible to consistently translate detailed features across views. First, features from the global texture are projected into each image plane of the viewpoint. The process initiates with ray casting, which essentially involves projecting rays from individual pixels of the 3D object to their respective spatial locations on the 2D image. This is followed by triplanar mapping, a procedure where the 2D texture map is accurately adapted to the three-dimensional geometry of the object, ensuring each point on the image correlates precisely with a point on the object's surface. The final step involves bilinear interpolation, a technique used to refine the texture of the object, effectively smoothing and harmonizing the texture details to eliminate any discontinuities or anomalies (Figure \ref{fig:viewpoints}). This process transforms three-dimensional objects into two-dimensional images. To encourage consistency, the model enforces a lighting-invariant view-consistency loss between pairs of translated frames. A detailed description of this loss function is presented in Section \ref{sec:Objective Functions}. By storing global texture information and enforcing view consistency, the model can generate videos that maintain consistent details over long duration rather than just between adjacent frames.

\section{Evaluation matrices}
A range of quantitative evaluation metrics have been proposed in literature to assess the performance of translation-based video synthesis models. In this section, we introduce and describe a comprehensive set of evaluation metrics commonly employed in the context of TVS tasks. Here, TP, TN, FP, FN are True Positive, True Negative, False Positive, and False negative, respectively, unless otherwise specified. We categorize these metrics based on the type of attributes they measure.
\subsection{Statistical Similarity Metrics}
These metrics quantify the statistical similarity between synthesized and real videos by comparing their feature distributions.

Fréchet Inception Distance (FID)  \cite{heusel2017gans}: 
FID measures the statistical divergence between feature distributions of real and synthesized frames extracted using a pretrained neural network. Lower FID indicates greater similarity.

\begin{gather}
FID = || \mu_1 - \mu_2||^2 + Tr(\sigma_1 + \sigma_2 -2\sqrt{(\sigma_1\sigma_2)})
\end{gather}

Here, $\mu_1$ and $\mu_2$ are the mean feature vectors of the real and generated video frames, respectively. $\sigma_1$ and $\sigma_2$ are the covariance matrices of the feature vectors for the real and the generated video frames, respectively. These matrices describe the spread or dispersion of the features in each set. $Tr(.)$ is the trace operator, which calculates the sum of the diagonal elements of a matrix, and $||.||^2$ is the squared Euclidean distance between the mean feature vectors.

Peak Signal-to-Noise Ratio (PSNR) \cite{wang2002universal}: 
PSNR computes pixel-level errors between synthesized and real frames. Higher PSNR signifies better reconstruction quality.

 \begin{gather}
PSNR = 10 log_{10} (\frac{R^2}{MSE})
 \end{gather}

Here, $R$ is the maximum possible pixel value in the image or video (typically 255 for 8-bit images), and $MSE$ is mean squared error, a measure of the average squared difference between corresponding pixels in the predicted and the ground-truth images or videos.

Structural Similarity Index (SSIM) \cite{wang2004image}:
SSIM is used to measure the similarity between two images. SSIM considers changes in structural information, luminance, and contrast of the image to provide accurate and comprehend similarity metric. SSIM value is a decimal between -1 and 1. A value of 1 indicates perfect similarity.
\begin{gather}
    \text{SSIM}(x, y) = \frac{(2\mu_x \mu_y + C_1)(2\sigma_{xy} + C_2)}{(\mu_x^2 + \mu_y^2 + C_1)(\sigma_x^2 + \sigma_y^2 + C_2)}
\end{gather}
Here, $x$ and $y$ are ground-truth and generated images, respectively. $\mu_x$, $\mu_y$ are the average of $x$ and $y$. $\sigma_x^2$, $\sigma_y^2$ are the variance and $\sigma_{xy}$ is the covariance of $x$ and $y$. $C_1$ and $C_2$ are constants used to stabilize the division.

\subsection{Semantic Consistency Metrics}
These metrics evaluate synthesized videos for consistency of semantic content objects, motions, and scenes.

Region Similarity (RS) \cite{wei2018video}:
RS quantifies semantic overlap between segmentation maps of synthesized and real frames. Higher values indicate better segmentation accuracy for video to label segmentation.
\begin{gather}
    RS = \frac{2TP }{2(TP + FP + FN)}
\end{gather}
In this framework, `True Positives' (TP), `False Positives' (FP), and `False Negatives' (FN) represent standard metrics used to evaluate the performance of video segmentation against the ground-truth mask.

Temporal Stability \cite{wei2018video}:
 Measures variability of semantic similarity scores across consecutive synthesized frames. Lower variability signifies more stable semantics.
\begin{gather}
    Temporal stability = 1 - Var(Dice)
\end{gather}
Here, Dice coefficient is a metric used to quantify the spatial overlap between two sets. It is often used in image segmentation tasks to measure the similarity between predicted and ground-truth regions. The Dice coefficient is calculated by $Dice =\frac{2TP}{2TP + FP + FN}$ and $Var(.)$ represents the variance of a quantity.

Pixel Accuracy (PA) \cite{long2015fully}: 
This metric assesses the accuracy of segmentation predictions in video synthesis. It measures the accuracy of pixel-level image segmentation by calculating the ratio of correctly classified pixels to the total number of pixels.
\begin{gather}
    Pixel Accuracy = \frac{TP + TN}{TP + TN + FP + FN}
\end{gather}

This metric is utilized specifically for assessing the performance in video-to-label and label-to-video tasks, providing a quantitative measure of the accuracy at the pixel level in these conversions.

Mean Pixel Accuracy (MPA) \cite{chen2019mocycle}: 
MPA computes the ratio of correct pixels on a per-class basis and averages them over all classes, providing insights into segmentation accuracy.
\begin{gather}
    Mean Pixel Accuracy = \frac{1}{C}\sum(PA per Class)
\end{gather}

Here, $C$ stands for the total number of classes or categories in the segmentation task. Chen et al. \cite{chen2019mocycle} utilized the `Mean Pixel Accuracy' metric for the evaluation of video-to-label and label-to-video tasks.

Intersection over Union (IoU) \cite{ren2015faster}: 
IoU calculates semantic overlap between predicted and ground-truth segmentation. Higher IoU values indicate better overlap.
\begin{gather}
    IoU = \frac{TP}{TP + FP + FN}
\end{gather}
In this framework, `True Positives' (TP), `False Positives' (FP), and `False Negatives' (FN) represent standard metrics used to evaluate the performance of video segmentation against the ground-truth mask for video-label segmentation tasks.

Mean Intersection over Union (mIoU) \cite{long2015fully}: 
The mIoU is then calculated as the mean of the IoU values across all classes. 
\begin{gather}
     mIoU = \frac{1}{c}\sum^C_{i=1}IoU_i
\end{gather}
Here, $C$ is the total number of distinct classes in the dataset.

Frequency Weighted Intersection-Over-Union (fwIoU) \cite{park2019preserving}: 
This metric evaluates semantic segmentation quality using a weighted intersection-over-union metric.
\begin{gather}
    fwIoU = \sum(IoU per Class) * (Class Frequency)
\end{gather}

Normalized FCN Score \cite{bansal2018recycle}:
FCN refers to a Fully Convolutional Network, which is a semantic segmentation model. This metric evaluates the quality of synthetic images by assessing their similarity to real images using a pre-trained fully convolutional network (FCN). The idea is to compare how similarly the network segments these two types of images. FCN processes both the real and the synthesized images to produce their respective segmentation maps. The quality of the synthesized images is then assessed based on how closely their segmentation maps match those of the real images. The Normalized FCN Score is essentially calculated by taking the average squared Euclidean differences between the real and the synthetic images and normalizing them by the squared Euclidean norm of the real images. A higher normalized FCN score indicates that the synthesized images look more realistic and closer to real images.
\begin{gather}
NormalizedFCN = 1 - \frac{1}{N}\sum^N_{i = 1}\frac{|| I_i - I_{i_{synthesized}}||^2}{|| I_i||^2}
\end{gather}

Here, $N$ represents the total number of frames being evaluated, $I_i$ refers to an individual real (ground-truth) frame from the dataset, and $I_{i_{synthesized}}$ corresponds to the synthetic frame generated by the model. $|| I_i - I_{i_{synthesized}}||^2$ calculates the squared Euclidean distance ($L^2-norm$) between the real frame ($I_i$) and its synthetic counterpart $I_{i_{synthesized}}$ . This measures the pixel-wise difference between the two frames. $|| I_i||^2$ represents the squared Euclidean norm of the real frame ($I_i$). It measures the total pixel energy or intensity in the real video.
In the labels-to-video translation task, semantic segmentation labels are fed as input to the generator to synthesize corresponding realistic-looking videos. The normalized FCN score allows a quantitative evaluation of how realistic the generated images are.

Average Content Distance for Identity (ACD-I) \cite{zhao2020towards}:
ACD-I is a specialized evaluation metric used primarily in the context of facial expression synthesis and manipulation, particularly in scenarios where maintaining the identity of the subject is crucial while altering their facial expressions. ACD-I metric quantifies the distance between the synthesized facial image and the original image in terms of content or identity features. ACD-I score is calculated as the average Euclidean distance between the real and synthetic feature vectors across a set of images. These feature vectors are extracted using pre-trained deep Convolutional Neural Networks (CNNs) and are specifically tuned to capture identity-related features such as eyes, nose, and mouth positions. Rather than using the final output layer, intermediate layers of these networks are used to extract feature vectors. These layers capture complex patterns and attributes of the images, such as textures, shapes, and contours, which are more relevant for identifying and preserving identity.  A lower ACD-I score indicates that the synthesized images closely retain the identity features of the original images.
\begin{gather}
    \text{ACD-I} = \frac{1}{N} \sum_{i=1}^{N} \sqrt{\sum_{j=1}^{M} (f_{i,j}^{\text{orig}} - f_{i,j}^{\text{synth}})^2}
\end{gather}

Here, $f_{i,j}^{\text{orig}}$ and $f_{i,j}^{\text{synth}}$ are the $j^{th}$ features of the $i^{th}$ image in the original and synthesized datasets, respectively. $\sqrt{\sum_{j=1}^{M} (f_{i,j}^{\text{orig}} - f_{i,j}^{\text{synth}})^2} $calculates the Euclidean distance between the feature vectors of the original (orig) and synthesized (synth) images. $M$ represents the number of features in the feature vector of image ($i$) and $N$ is the total number of images in the dataset.

\subsection{Motion Consistency Metrics}
These metrics assess how consistently motion characteristics are preserved in the synthesized videos.

Pose Error \cite{chan2019everybody}:
This approach quantifies discrepancies between the estimated pose $P^{'}$ and the ground-truth pose $P$ across different frames. A lower error value is indicative of a more precise pose translation. The first step is to encode the body pose of each person in the video frames using OpenPose \cite{cao2017realtime}. OpenPose estimates the 2D positions of various joints (key points) on the human body, such as elbows, knees, and shoulders. These key points are then used to create a skeleton by connecting the joints with lines. The pose error between the ground-truth ($P$) and generated pose ($P^{'}$), each consisting of $n$ joints, is then calculated by summing up the Euclidean distances for each pair of corresponding joints and normalizing this sum by the number of joints ($n$). This is essentially measuring the straight-line distance between the two points in 2D space.

\begin{gather}
    PoseError = \frac{1}{n} \sum_{k=1}^{n} \| P_k - P'_k \|^2
\end{gather}
Here, $\| P_k - P'_k \|^2$ calculates the Euclidean distance between the k-th joint in the estimated pose $P^{'}$ and the ground-truth pose $P$. The symbol $\sum$ is used to sum up the distances across all joints in the pose.

Warping Error \cite{wang2022learning}:
This is an evaluation metric used to quantify the temporal consistency of video sequences. It measures how well the warping or alignment process has been performed between consecutive frames in a video. This metric is particularly relevant in video-to-video translation tasks, where maintaining the coherence of object motions across frames is crucial for producing realistic and visually pleasing results.
\begin{gather}
    F^{\prime}_t = Warp(F_{t-1},F_{t-2})
\end{gather}

\begin{gather}
    Warping Error = \frac{1}{W * H * T}\sum || F_t - F^{\prime}_t ||
\end{gather}

Here, $Warp()$ function generates the frame $F^{\prime}_t$ based on the temporal information of the previous two frames ($t-1, t-2$). $Warping Error$ calculates the error between the warped frame $F^{\prime}_t$ and the ground-truth frame $F_t$.
$\frac{1}{W * H * T}$ normalizes the error based on the dimensions of the frames and the number of frames in the video. $W$ represents the width of the frames, $H$ represents the height, and $T$ represents the total number of frames in the video sequence. This normalization ensures that the metric is not influenced by the size or length of the video. $\sum || F_t - F^{'}_t ||$ calculates the pixel-wise Euclidean distance ($L^2-norm$) between the original frame $F_t$ and the corresponding warped frame $F^{'}_t$ at each pixel location. It measures how much the pixel values have changed during the warping process.

\subsection{Subjective Matrices}
These metrics measure perceptual realism, smoothness, and naturalness of synthesized videos via human evaluations.

\quad Human Subjective Score (HSS) \cite{wang2019few}:
Ratings collected from user studies judging quality of synthesized videos. Higher is better.
\begin{gather}
    HSS = \frac{1}{N}\sum(User Scores)
\end{gather}

Here, $N$ represents the total number of participants or users involved in the study.

\section{Objective Functions}
\label{sec:Objective Functions}
Several state-of-the-art research papers have introduced novel loss functions that are essential to attaining outstanding outcomes in the field of video synthesis and translation. Advanced generative models are trained using these loss functions as the guiding principles, resulting in stunning transformations and retaining temporal consistency. We gain important insights into the fundamentals of cutting-edge video synthesis methods by examining the nuances of these loss functions, paving the way for more convincing and realistic video transformations.
\subsection{Adversarial Objectives}

\quad Adversarial Loss \cite{goodfellow2020generative}:
The adversarial loss encourages the generator networks GX and GY for domains $X$ and $Y$, respectively to produce outputs that are indistinguishable from real data, according to the discriminators DX and DY. This loss is inspired by GANs and helps in generating realistic samples. Here, $x_t$  represents a frame in domain $X$ and $y_s$ represents a frame in domain $Y$.

For GX: 
\begin{gather}
    \genfrac{}{}{0pt}{}{min}{GX}\genfrac{}{}{0pt}{}{max}{DX} Lg(GX,DX) = \sum_t log DX(x_t) + \sum_s log(1-DX(GX(y_s)))
\end{gather}

For GY:
\begin{gather}
    \genfrac{}{}{0pt}{}{min}{GY}\genfrac{}{}{0pt}{}{max}{DY} Lg(GY,DY) = \sum_s log DY(y_s) + \sum_t log(1-DY(GY(x_t)))
\end{gather}

Discriminator Loss \cite{goodfellow2020generative}:
It distinguish between the real and generated frames. $DX$, $DY$ are the discriminators for domains $X$ and $Y$, respectively.

\begin{gather}
    Lc(GX,GY) = \sum || x_t - GX(GY(x_t))||^2
\end{gather}

\subsection{Temporal Consistency Objectives}

\quad Recurrent Loss \cite{bansal2018recycle}:
The recurrent loss trains the temporal predictors PX and PY for domains $X$ and $Y$, respectively, to predict future samples based on the previous two samples in a sequence. This loss helps the model capture temporal dependencies in the data.
\begin{gather}
    L_{\tau}(PX) = \sum || x_{t+1} - PX(x_{1:t})||^2
\end{gather}
Recycle Loss \cite{bansal2018recycle}:
The recycle loss combines the concepts of cycle consistency and temporal prediction, ensuring that sequences map back to themselves across domains and time.

\begin{gather}
    L_r(GX,GY,PY) = \sum || X_{t+1} - GX(PX(GY(x_{1:t})))||^2
\end{gather}

Tendency-Invariant loss \cite{liu2020unsupervised}:
The loss is formulated based on the principle that the prevailing trends in frame modifications should persist in predicted frames, thereby enhancing the preservation of source domain content and elevating output quality. This innovative loss function operates as follows. First, it takes two input frames, $x_t$ and $x_{t+1}$, representing the current and subsequent frames, respectively. Subsequently, $PX(x_{t-1}^t)$ is employed to predict $x_{t+1}$, drawing upon information from $x_{t-1}$ and $x_t$. The loss is calculated as the disparity between two L1 losses, revealing the tendency or direction of frame modifications rather than the modifications themselves. By minimizing the tendency-invariant loss, the framework ensures that predicted frames align with the observed trends in frame modifications, resulting in a more effective preservation of content and improved overall output quality.
\begin{gather}
    \genfrac{}{}{0pt}{}{min}{PX} L_{inv}(PX) = \sum_t || ((PX(x_{t}^{t+1}) - PX(x_{t-1}^t)) - (PX(x_{t-1}^t) - x_t))
    - ((x_{t+2} - x_{t+1}) - (x_{t+1}-x_t))||_1
\end{gather}

Smoothness Regularization Loss \cite{fan2019controllable}:
The regularization technique aims to reduce the difference between adjacent video frames generated by the network. This regularization serves a dual purpose: preventing mode-dropping issues and enhancing the temporal smoothness of the generated video clips. The regularization term, denoted as $Rt$, is defined as the sum of two $L1-norms$, each calculated between a generated video frame $V(a)$ and its neighboring frames $V(a - \Delta a)$ and $V(a + \Delta a)$, respectively,where $\Delta a$ represents a small increment in the video sequence. The method can be defined as follows:
\begin{gather}
    Rt = || V(a) - V(a - \Delta a)||_1 + || V(a) - V(a + \Delta a)||_1
\end{gather}

View-Consistency Loss \cite{rivoir2021long}:
This loss is a specialized loss function used to ensure consistency between two views of a simulated scene. Specifically, two random views $i$ and $j$ of the same simulated scene are sampled and translated. View $j$ is then warped into the pixel space of view $i$. The loss function is then applied between the warped view $j$ and view $i$ to encourage consistency. The loss computes the angle between RGB vectors at each pixel instead of using a standard L1 or L2 loss. This angle loss is defined as:
\begin{gather}
        L_{vc} = \left( \frac{1}{|M|} \right) \sum_{(x,y)} \cos^{-1} \left( \frac{\mathbf{b}_{i}^{xy} \cdot \mathbf{w}_i(\mathbf{b}_j)^{xy}}{||\mathbf{b}_i^{xy}|| \, ||\mathbf{w}_i(\mathbf{b}_j)^{xy}||} \right)
\end{gather}

Here, M is the set of valid matching pixel locations. $b_i^{xy}$ is the RGB vector at this location $(x,y)$, $w_i(.)$ is the warping operator into $b_i$'s pixel space. 
Note: Angle between $u$ and $v$ can be computed by $cos^{-1}((u.v)/(||u||||v||))$. Consistency hue is enforced in corresponding locations while allowing varying brightness.

\subsection{Content Preservation Objectives}

\quad Emotion Loss \cite{shen2019facial}:
This loss function uses a pre-trained classifier to predict the emotional labels corresponding to each generated frame. This loss function is used for applications including changing facial expression from sad to smiling, angry to happy, etc. A logit ($z$) is generated from the final layer of the classifier through a softmax function to compute the likelihood of each emotion class ``$e\epsilon E$''. The likelihood is computed as the exponential of ``$z_e$'' divided by the sum of the exponentials of all classes.

\begin{gather}
g_e = \frac{exp(z_e)}{\sum_{e\epsilon E} exp(z_e)}
\end{gather}
The loss was obtained by minimizing the deviation between the desired emotion class likelihood, denoted as ``$g_e$'' and the maximum possible likelihood (1) for the total number of frames ($N$) in the sequence.
\begin{gather}
    L_{emotion} = \frac{1}{N} \sum_{i=1}^N 1 - g_e^i
\end{gather}
The objective aims to align the predicted emotions with the desired emotional content, enhancing the emotional relevance and quality of the generated frames.

Ranking Loss \cite{zhao2020towards}:
The key idea of ranking loss is that the distance between the refined video $V''_t$ and the target ground-truth video $V_t$ should be smaller than the distance between $V''_t$ and the input video $V'_t$. Optimizing this ranking loss encourages the output to be closer to the ground-truth in terms of motion features.
\begin{gather}
L_{\text{rank}}(V_t, V'_t, V''^{_t}) = - \log \left( \frac{e^{- ||f_{D_V}(V''_t) - f_{D_V}(V_t)||_1}}{e^{-||f_{D_V}(V''_t) - F{D_V}(V_t)||_1} + e^{-||f_{D_V}(V''_t) - f_{D_V}(V'_t)||_1}} \right)
\end{gather}

Here, $f_{D_V}$ is a differential function to compute the motion feature using video discriminator $D_V$.

\section{Dataset}
A range of datasets have been used to advance TVS research by providing diverse domains for training and benchmarking models. These datasets can be categorized based on supporting image-to-video (i2v) or video-to-video (v2v) translation tasks.

For i2v translation, the majority of datasets focus on modeling facial expressions and emotions. The CK+ \cite{lucey2010extended} dataset offers rich annotations of facial landmarks in videos displaying emotions, enabling in-depth emotion analysis. Its extension CK++ \cite{fan2019controllable} adds color videos to boost diversity. The Cheeks and Eyes Dataset \cite{shen2019facial} provides more nuanced expressions involving eyes and cheeks to complement CK+. 

CK+ and Cheeks and Eyes are limited in subjects and domains. The larger MUG Facial Expression dataset \cite{aifanti2010mug} features 86 subjects across a balanced gender distribution, capturing anger, fear, disgust, happiness, sadness, and surprise. It provides more extensive diversity. However, these datasets are still restricted to controlled facial videos.

In contrast, v2v translation involves more varied real-world domains. Cityscapes \cite{cordts2015cityscapes} and Apolloscape \cite{huang2018apolloscape} provide street scenes across diverse cities, with semantic segmentation labels to learn spatial understanding. DAVIS 2017 \cite{pont20172017} focuses on high-quality videos of foreground objects for semantic segmentation tasks. Viper \cite{richter2017playing} contains realistic computer game videos across different environments. These facilitate translating visual aspects such as scene layouts and objects.

Medical imaging poses unique v2v challenges such as medical image segmentation, laparoscopic image dataset \cite{rivoir2021long} for generating laparoscopic images from their corresponding masks, conversion of CT image to MRI image, etc. MRCT \cite{akkus2017predicting, vallieres2017radiomics} consists of MR and CT scan volumes, enabling translation between modalities. Volumetric MNIST \cite{lecun1998mnist} simulates a 3D scanning domain for moving digit ($0-9$) videos. Both require translating 3D spatio-temporal data.

For human actions, Human3.6M \cite{ionescu2013human3} and Penn Action \cite{yang2018pose} provide poses and joints in videos of daily activities. These are crucial for modeling complex motions and poses. GTA Segmentation \cite{srivastava2015unsupervised} adds gameplay videos to introduce man-made environments.

Certain datasets assess specific capabilities. Face-to-Face \cite{bansal2018recycle} and Flower-to-Flower \cite{bansal2018recycle} evaluate style and content translation robustness. Sunrise \& Sunset \cite{bansal2018recycle} examines alignment between similar scenes. These test generalization to limited, focused domains.

In summary, facial translation-focused video datasets allow isolating facial motions but lack diversity. Large-scale scene datasets offer more variability at the cost of increased complexity. Medical and human action datasets demand translating specialized 3D and pose structures. Together, they provide complementary domains to further TVS progress. 

\section{Quantitative Analysis}

This section presents a comprehensive quantitative evaluation of various approaches to TVS. The effectiveness of these approaches is heavily influenced by the chosen methodology and datasets. In the video-to-label translation task (Video2Label in Table \ref{tab:qualitative approach}), the temporal constraints based method achieved a mIoU of 0.095 \cite{bansal2018recycle} whereas the optical flow-based method obtained an mIoU between 0.091 \cite{park2019preserving} and 0.14 \cite{wang2022learning} using Viper dataset. The reason for the superior performance of \cite{wang2022learning} is the introduction of a synthetic optical flow to represent flawless motion across domains, while the earlier methods used the previous two frames to generate the current frame, causing incorrect motion estimations. For a similar task, paired v2v obtained a mIoU of 0.41 \cite{wang2019few} using Cityscapes dataset. The high performance of the Paired v2v method is attributed to its direct frame-to-frame correspondence, enhancing the accuracy and consistency of segmentation along with pixel-level annotation of Cityscapes dataset. In the label-to-video translation task (Label2Video in Table \ref{tab:qualitative approach}), paired v2v achieved an FID score between 49.89 \cite{mallya2020world} and 144.2 \cite{wang2019few} on Cityscapes dataset. The superior performance by  \cite{mallya2020world} is attributed to its consideration of all the previous frames to construct a ``guidance image'' to generate the current frame, which provides more solid spatio-temporal information than considering the previous one or two frames as considered by other approaches. Similar to video-to-label translation task, \cite{wang2022learning} shows superior performance on label-to-video translation task with mIoU of 0.11 \cite{wang2022learning} compared to mIoU of 0.07 \cite{bansal2018recycle}, and 0.08 \cite{park2019preserving} on Viper dataset. In the context of medical image segmentation, extended i2i shows superior performance on laparoscopic image generation from their corresponding mask with an FID score of 26.8 \cite{rivoir2021long} over temporal predictor-based method with an FID score of 61.5 \cite{bansal2018recycle}. The introduction of neural rendering on top of the standard image-to-image classification model contributes to better generating the spatial information of the frames and maintaining the temporal consistency over a longer sequence.
In the realm of facial expression detection (Facial Expression in Table \ref{tab:qualitative approach}), the image-to-video (i2v) model demonstrated an Average Content Distance for Identity (ACD-I) score of 0.18 \cite{zhao2020towards} whereas RNN based approach provided an ACD-I of 0.29 \cite{tulyakov2018mocogan} on MUG dataset. The introduction of a two-stage neural network for facial expression retargeting and human pose forecasting is capable of capturing foreground information and smothering facial transition than a regular RNN-based framework which considers all the pixels in the image. An introduction of ``tendency-invariant loss'' improves the performance of the temporal constraints-based method with an FID score of 49.5 \cite{liu2020unsupervised} for face-to-face transition (Face2Face Transition in Table \ref{tab:qualitative approach}) and an FID score of 112.2 \cite{liu2020unsupervised} for flower-to-flower transition (Flower2Flower in Table \ref{tab:qualitative approach}) compared to Recycle-GAN with an FID score of 53.4 \cite{bansal2018recycle} and 118.3 \cite{bansal2018recycle} for face-to-face and flower-to-flower transition, respectively.

Our analysis indicates that methods that incorporate more contextual information and advanced motion representation tend to perform better. Furthermore, enhancing traditional models with neural rendering techniques can lead to substantial improvements in temporal consistency and spatial details, especially in complex tasks such as medical image generation and facial expression detection.

\begin{table*}[h]
\small
\caption{Quantitative analysis of different approaches.}
\label{tab:qualitative approach}
\begin{tabular}{|p{1.3cm}|p{1.5cm}|p{1.5cm}|p{1.5cm}|p{1.4cm}|p{1.5cm}|p{1.7cm}|p{1.0cm}|p{1.3cm}|}
\hline
Tasks & \multicolumn{1}{c|}{Datasets} & \multicolumn{7}{c|}{Methods [Metrics: Score [Reference]]} \\ \hline
&   & i2v & Paired v2v & 3D GAN & Temporal Constraints & Optical Flow & RNN & Extended i2i \\ \hline
& Viper & & & & mIoU:0.095\cite{bansal2018recycle} & mIoU:0.091\cite{park2019preserving}, 0.14 \cite{wang2022learning} & & \\ \cline{2-9}
Video2Label & GTA & & & PA:0.60\cite{bashkirova2018unsupervised} & & & & \\ \cline{2-9}
& Cityscapes & & mIoU:0.41\cite{wang2019few} & & & & & \\ \hline

Label2Video & GTA & & & HSS:0.68\cite{bashkirova2018unsupervised} & & & & \\ \cline{2-9}
& Cityscapes & & FID:144.2\cite{wang2019few}, 69.07 \cite{wang2018video}, 49.89 \cite{mallya2020world} & & & & & \\ \cline{2-9}
& Viper & & & & mIoU:0.07\cite{bansal2018recycle} & mIoU:0.08\cite{park2019preserving}, 0.11 \cite{wang2022learning} & & \\ \cline{2-9}

&Laparoscopic Image Dataset  & & & & FID:61.5\cite{bansal2018recycle} & & & FID:26.8\cite{rivoir2021long} \\ \hline

Facial Expression & Cheeks\&Eyes & ACD-I: 0.31\cite{shen2019facial}& & & & & & \\ \cline{2-9}
& CK+, CK++ & SSIM:0.95\cite{fan2019controllable} & & & & & & \\ \cline{2-9}
& MUG & ACD-I: 
0.18\cite{zhao2020towards}& & & & & ACD-I: 0.29\cite{tulyakov2018mocogan}& \\ \hline
Face2Face Transition & Face2Face & & & &  FID:53.4\cite{bansal2018recycle}, 49.5\cite{liu2020unsupervised} & & & \\ \hline
Flower2 
 Flower& Flower2 Flower& & & & FID:118.3\cite{bansal2018recycle}, 112.2\cite{liu2020unsupervised} & & & \\ \hline
VIPER2 Cityscapes & VIPER Cityscapes & & & mIoU:0.25\cite{bansal2018recycle} & mIoU:0.35\cite{park2019preserving} & & & \\ \hline
Video Impainting & DAVIS 2017 & & & & & & & FID:17.8\cite{szeto2021hypercon} \\ \hline
\end{tabular}
\footnotesize{Abbreviations: \textbf{MPA}: mean pixel accuracy, \textbf{mIoU}: mean intersection over union, \textbf{PA}: pixel accuracy, \textbf{HSS}: human subjective score, \textbf{FID}: Fréchet inception distance, \textbf{FCN}: normalized FCN score, \textbf{ACD-I}: average content distance for identity, \textbf{SSIM}: structural similarity index}
\end{table*}

\section{Discussion}
This survey provides a comprehensive overview of the emerging field of translation-based video-to-video synthesis (TVS). Through a systematic analysis of current state-of-the-art techniques, several key insights can be gleaned regarding the progress, remaining challenges, and future directions for TVS research.

A significant trend observed is the shift from early paired v2v methods relying on explicit frame correspondences between input and output videos, towards more flexible unpaired v2v approaches that learn implicit relationships between different video domains. Unpaired methods offer greater scalability by avoiding exhaustive frame-level annotations. However, unpaired v2v translation remains an inherently difficult undertaking, necessitating complex constraints and objectives to align content across domains while maintaining spatio-temporal coherence.

The survey reveals a diversity of strategies employed to address this challenge. Architecturally, 3D GANs directly model videos as spatio-temporal data. Temporal consistency methods introduce specialized objectives such as future frame prediction and tendency-invariant losses. Optical flow techniques focus on translating motion dynamics. RNN models exploit long-range temporal contexts. Extending image translators to video remains an attractive option for its simplicity, but risks propagating errors on a per-frame basis.

Each strategy exhibits characteristic strengths and limitations. 3D GANs struggle to align content semantics across domains. Temporal consistency approaches may fail to capture nuanced frame evolutions. Inaccurate optical flow estimation can undermine motion translation. RNNs are computationally intensive. Image extension methods lack mechanisms for long-term consistency. Evidently, there remains ample scope for devising more robust and integrated solutions.

The survey also highlights the value of multi-stage and residual learning strategies for high-quality video synthesis. Many current methods decompose the task into coarse translation followed by refinement, avoiding directly regressing details in one single stage. These concepts could inform future architectures.

One of the core takeaways from this study is the critical role of diverse datasets in advancing the field of video-to-video synthesis. Researchers have at their disposal a rich array of datasets, such as Cityscapes, Apolloscape, DAVIS 2017, and many others, that serve as benchmarks for evaluating the capabilities of translation models. These datasets encompass a wide range of applications, from urban scenes to medical imaging, enabling the evaluation of algorithms in various domains.

The future holds great promise for the developing field of video-to-video (v2v) synthesis. The possibility of generating more convincing, solid, and user-controlled video transformations grows more real as these techniques develop. For v2v
approaches, the pursuit of long-term temporal consistency and the faithful depiction of 3D world dynamics remains
an under-explored area where further study is anticipated to result in significant progress. The development of new
architectures and regularization methods holds a promise in improving the realism of synthesized videos and
bridging the gap between artificially created content and real content. A growing area of interest is the interplay
between the image-to-video (i2v) and video-to-video (v2v) approaches. Combining these paradigms has the potential to
produce comprehensive systems that offer both fine-grained user control and realistic, coherent video creation. These
initiatives are expected to expand the horizon of content creation and manipulation and offer unique and limitless
opportunities.

\section{Conclusion}
This survey provides a structured taxonomy of translation-based video-to-video synthesis (TVS) literature. We categorize methods into image-to-video, paired video-to-video, and unpaired video-to-video types. Architecturally, we identify five key strategies: 3D GANs, temporal constraint-based, optical flow, RNN, and extended image translators. Through detailed analysis, we compare approaches and highlight limitations regarding spatio-temporal consistency and realism. Our survey includes extensive benchmarking using reported quantitative results. We discuss diverse real-world datasets and identify complementary domains for future research. Additionally, we analyze loss functions for realism, consistency, and content preservation. We suggest multi-stage learning and new evaluation metrics as promising directions. Overall, this survey delivers a comprehensive analysis of TVS literature and provides valuable insights to advance this emerging field.

\bibliographystyle{unsrt} 
\bibliography{references} 

\end{document}